\documentclass[11pt]{article}

\PassOptionsToPackage{table,dvipsnames}{xcolor}
\PassOptionsToPackage{numbers,sort&compress}{natbib}

\usepackage[T1]{fontenc}
\usepackage[utf8]{inputenc}
\usepackage{url}
\usepackage{microtype}
\usepackage{times}
\usepackage{xcolor}
\usepackage{graphicx}
\usepackage{grffile}
\usepackage{svg}
\usepackage{amsmath}
\usepackage{amssymb}
\usepackage{algorithm}
\usepackage{algpseudocode}
\usepackage{booktabs}
\usepackage{multirow}
\usepackage{colortbl}
\usepackage{makecell}
\usepackage{placeins}
\usepackage{float}
\usepackage{xspace}

\newcommand{\perceptronlogo}[1][0.26\linewidth]{\includegraphics[width=#1]{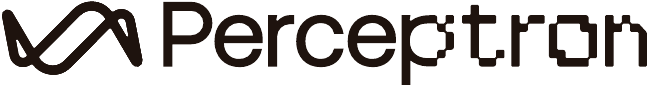}}

\usepackage{perception_conference}
\iclrfinalcopy
\title{SkillRater: Untangling Capabilities in Multimodal Data}
\author{Naveen Sahi \qquad Jeremy Dohmann \qquad Armen Aghajanyan \qquad Akshat Shrivastava}
\titlegraphic{\perceptronlogo[0.30\linewidth]}
\date{\today}

\newcommand{\systemname}{\textsc{SkillRater}\xspace}

\definecolor{eqParamBg}{RGB}{226,238,255}
\definecolor{eqStateBg}{RGB}{226,245,245}
\definecolor{eqDataBg}{RGB}{230,243,229}
\definecolor{eqWeightBg}{RGB}{255,238,225}
\definecolor{eqLossBg}{RGB}{255,248,220}
\definecolor{eqStepBg}{RGB}{245,236,255}
\definecolor{eqObjBg}{RGB}{242,233,255}
\definecolor{eqLabelBlue}{RGB}{24,96,171}
\definecolor{eqLabelGreen}{RGB}{74,132,58}
\definecolor{eqLabelPurple}{RGB}{132,66,174}
\definecolor{eqLabelOrange}{RGB}{201,101,39}
\definecolor{eqLabelGold}{RGB}{153,110,24}

\newcommand{\eqbg}[2]{%
  \begingroup
  \setlength{\fboxsep}{1pt}%
  \mathchoice
    {\colorbox{#1}{$\displaystyle #2$}}
    {\colorbox{#1}{$\textstyle #2$}}
    {\colorbox{#1}{$\scriptstyle #2$}}
    {\colorbox{#1}{$\scriptscriptstyle #2$}}%
  \endgroup
}

\begin{document}
\raggedbottom
\maketitle

\begin{abstract}
Data curation methods typically assign samples a single quality score. We argue this scalar framing is fundamentally limited: when training requires multiple distinct capabilities, a monolithic scorer cannot maximize useful signals for all of them simultaneously. Quality is better understood as multidimensional, with each dimension corresponding to a capability the model must acquire. We introduce \systemname, a framework that decomposes data filtering into specialized raters—one per capability, each trained via meta-learning on a disjoint validation objective—and composes their scores through a progressive selection rule: at each training stage, a sample is retained if any rater ranks it above a threshold that tightens over time, preserving diversity early while concentrating on high-value samples late. We validate this approach on vision language models, decomposing quality into three capability dimensions: visual understanding, OCR, and STEM reasoning. At 2B parameters, \systemname improves over unfiltered baselines by 5.63\% on visual understanding, 2.00\% on OCR, and 3.53\% on STEM on held out benchmarks. The learned rater signals are near orthogonal, confirming that the decomposition captures genuinely independent quality dimensions and explaining why it outperforms both unfiltered training and monolithic learned filtering.
\end{abstract}

\section{Introduction}

Most data filtering methods assign each training example a scalar quality score \cite{brown2020language}. This framing assumes quality is one dimensional. We show it is not: when training requires multiple distinct capabilities, a monolithic scorer cannot maximize useful signals for all of them simultaneously. Quality is better understood as multidimensional, with independent axes corresponding to the capabilities the model must acquire.

Existing approaches fail to capture this structure. CLIP score filtering \cite{schuhmann2022laion5b, gadre2024datacomp} measures image text alignment but does not reflect sample utility for reasoning tasks. Unsupervised clustering methods \cite{TODO_climb} discover natural data groupings but provide no mechanism to target specific capabilities. DataRater \cite{calian2025datarater}, a meta learned scoring function, achieves strong results for text but underperforms on multimodal data. We trace this failure to a structural bottleneck: the capabilities we care about have near orthogonal data requirements, so compressing them into one scalar induces tradeoffs that hurt overall performance.

We introduce \systemname, a framework that decomposes data filtering into specialized raters aligned to distinct capabilities. Each rater is trained via meta learning on a disjoint validation objective, producing scores that reflect sample utility for its assigned capability. A curriculum progressively tightens selection thresholds over the union of rater scores, preserving diversity early while concentrating on high quality samples late. We validate this approach on vision language models at the 2B parameter scale \cite{perceptronai_isaac01_paper}, decomposing quality into three capability dimensions: visual understanding, OCR, and STEM reasoning.

On benchmarks never seen during rater training, \systemname improves over unfiltered baselines by 5.63\% on visual understanding, 2.00\% on OCR, and 3.53\% on STEM reasoning. Analysis of the learned raters confirms our thesis: the three scoring functions are near orthogonal (mean pairwise correlation 0.020, effective dimensionality 2.99 out of 3.0), demonstrating that the decomposition captures genuinely independent quality dimensions. Curriculum scheduling outperforms all static top k filtering strategies, and raters trained at 1B parameters transfer to 2B without retraining.

Our main contributions are:
\begin{itemize}
  \item \textbf{Quality is multidimensional.} We provide empirical evidence that scalar quality scoring is insufficient when training requires multiple capabilities. Decomposed raters produce near orthogonal scores (effective dimensionality 2.99 out of 3.0) and outperform monolithic scoring.
  \item \textbf{Capability aligned filtering.} We propose \systemname, a framework that trains specialized raters via meta learning on disjoint validation objectives, combined with curriculum scheduling over their union.
  \item \textbf{Generalization across benchmarks and scales.} On held out benchmarks, \systemname improves over unfiltered baselines by 2.0\% to 5.6\% across capability dimensions. Raters transfer from 1B to 2B parameters without retraining.
\end{itemize}

% Our results suggest that capability aligned decomposition is a more effective paradigm for multimodal data curation than monolithic quality scoring. Crucially, gains on held-out benchmarks demonstrate that \systemname learns transferable quality signals rather than overfitting to specific evaluation targets. The uncorrelated nature of our raters provides researchers with independent levers for adjusting the balance between capability domains. We release our framework to enable researchers to construct task-specific data regimes for their own multimodal training pipelines.

\section{Background: DataRater}
We build on the DataRater framework, which provides the bilevel optimization template we extend to multiple capabilities. 
DataRater \cite{calian2025datarater} is a bilevel meta-learning method that learns a rater $r_{\phi}$ assigning each training example $z$ a scalar weight $w=r_{\phi}(z)\in[0,1]$. The goal is to choose weights that improve performance on a held-out validation objective.
This bilevel formulation builds on a line of work that frames data selection 
as bilevel optimisation \cite{maclaurin2015gradient, lorraine2020optimizing, wang2020optimizingdata, ren2018learning}.

\paragraph{Bilevel objective.}
Let $\theta$ denote the parameters of the (vision-)language model being trained. DataRater can be written as
\begin{align}
\min_{\eqbg{eqParamBg}{\phi}}\; &
\underbrace{
\eqbg{eqObjBg}{\mathcal{L}_{\mathrm{val}}}
\Bigl(
\underbrace{\eqbg{eqStateBg}{\theta^{\star}(\phi)}}_{\textcolor{eqLabelBlue}{\text{post-inner model}}}
\Bigr)
}_{\textcolor{eqLabelPurple}{\text{validation objective}}} \\
\text{s.t.}\; &
\eqbg{eqStateBg}{\theta^{\star}(\phi)}=\arg\min_{\eqbg{eqStateBg}{\theta}}\;
\underbrace{
\mathbb{E}_{z\sim \eqbg{eqDataBg}{\mathcal{D}_{\mathrm{train}}}}
\Bigl[
\underbrace{\eqbg{eqWeightBg}{r_{\phi}(z)}}_{\textcolor{eqLabelOrange}{\text{sample weight}}}\,
\overbrace{\eqbg{eqLossBg}{\ell(\theta;z)}}^{\textcolor{eqLabelGold}{\text{per-sample loss}}}
\Bigr]
}_{\textcolor{eqLabelGreen}{\text{weighted training loss}}}.
\end{align}
In practice, $\theta^{\star}(\phi)$ is approximated by running a finite number of inner-loop SGD steps.

\paragraph{Inner-loop training with weighted gradients.}
In the inner loop, we initialize the base model parameters $\theta^{(0)}=\theta$ and run $S$ weighted SGD steps. For a minibatch $\mathcal{B}$ and learning rate $\alpha$,
\begin{equation}
\eqbg{eqStateBg}{\theta^{(s+1)}}\;=\;\eqbg{eqStateBg}{\theta^{(s)}}-
\underbrace{\eqbg{eqStepBg}{\alpha}}_{\textcolor{eqLabelPurple}{\text{inner step size}}}\,
\underbrace{
\nabla_{\eqbg{eqStateBg}{\theta^{(s)}}}
\left(
\frac{1}{|\eqbg{eqDataBg}{\mathcal{B}}|}
\sum_{z\in\eqbg{eqDataBg}{\mathcal{B}}}
\eqbg{eqWeightBg}{r_{\phi}(z)}\,
\eqbg{eqLossBg}{\ell(\theta^{(s)};z)}
\right)
}_{\textcolor{eqLabelGreen}{\text{weighted microbatch gradient}}},
\quad s\in\{0,\dots,S-1\}.
\end{equation}
The post-inner-loop state is denoted $\theta^{(S)}$.

\paragraph{Outer-loop meta-update.}
The rater parameters $\phi$ are updated to improve held-out performance on the post-inner-loop state:
\begin{equation}
\eqbg{eqParamBg}{\phi} \leftarrow \eqbg{eqParamBg}{\phi} -
\underbrace{\eqbg{eqStepBg}{\eta}}_{\textcolor{eqLabelPurple}{\text{outer step size}}}\,
\underbrace{\nabla_{\eqbg{eqParamBg}{\phi}}
\eqbg{eqObjBg}{\mathcal{L}_{\mathrm{val}}}
\bigl(\eqbg{eqStateBg}{\theta^{(S)}(\phi)}\bigr)}_{\textcolor{eqLabelBlue}{\text{meta-gradient on rater}}}.
\end{equation}
In its original form, DataRater uses this bilevel objective to learn a weighting function that prioritizes samples that most improve validation performance.

\paragraph{Using the trained rater.}
Once trained, $r_{\phi}$ can be used to filter (top-$k$) or reweight incoming training data, enabling learned data curation without hand-crafted heuristics.
This formulation assumes a single validation objective. When training requires multiple capabilities with competing data requirements, a single rater introduces cross capability tradeoffs.

\paragraph{Notation.}
We use $\theta^{(0)}$ for the initial inner-loop base model state, $\theta^{(s)}$ for inner step $s$, and $\theta^{(S)}$ for the post-inner-loop state after $S$ steps. $\phi$ denotes rater parameters. $z$ denotes one multimodal training example. $\mathcal{B}$ and $\mathcal{V}$ denote train and validation microbatches. $\alpha$ and $\eta$ are inner-loop and outer-loop learning rates.

\section{\systemname}
\begin{figure}[t]
  \centering
  \includegraphics[width=0.95\linewidth]{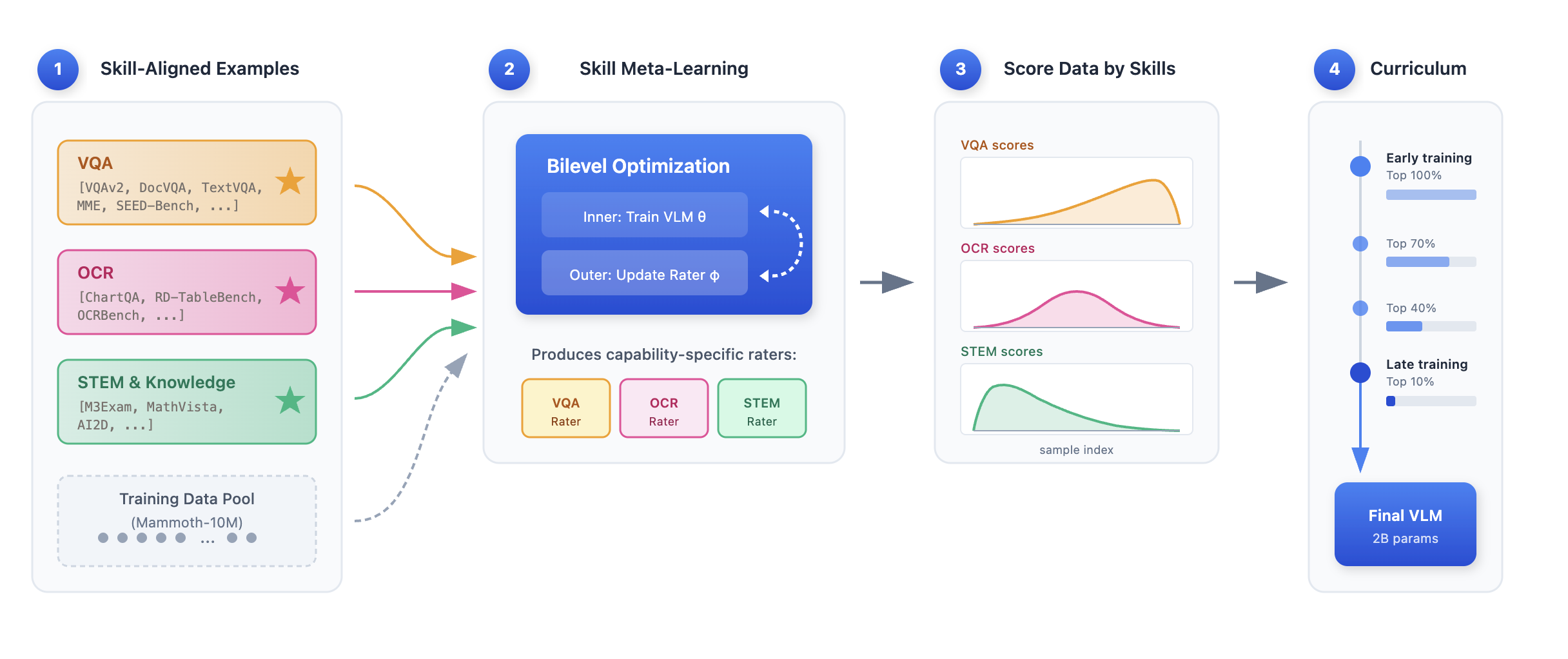}
  \caption{The \systemname pipeline. Capability aligned benchmarks and a training data pool are used in a bilevel meta-learning framework to learn per-capability raters (Visual Understanding, OCR, STEM). These raters score the training data, which is then filtered via a curriculum that progressively selects higher quality samples throughout training.}
  \label{fig:skillrater_overview}
\end{figure}

\subsection{Capability-Aligned Rater Decomposition}
We decompose filtering into $C$ capability-specific raters, each trained on a disjoint validation objective. For each capability $c \in \{1, \dots, C\}$, we solve:
\begin{align}
\min_{\phi_c}\; &\mathcal{L}_{\mathrm{val}}^{(c)}\bigl(\theta^{(S)}(\phi_c)\bigr) \label{eq:skillrater_outer}\\
\text{s.t.}\; &\theta^{(S)}(\phi_c) = \texttt{InnerLoop}\bigl(\theta, r_{\phi_c}, \mathcal{D}_{\mathrm{train}}\bigr), \notag
\end{align}
where $\mathcal{L}_{\mathrm{val}}^{(c)}$ is a capability-specific validation objective and $r_{\phi_c}$ is the corresponding rater. The inner loop follows the same bilevel structure as DataRater (Section~2), but each rater sees gradients only from its own capability benchmarks.

We instantiate $C{=}3$ raters with disjoint validation objectives (Table~\ref{tab:held_out_split}):
\begin{itemize}
  \item \textbf{Visual Understanding Rater:} Optimizes for performance on SEED-Bench~\cite{li2023seedbench}, A-OKVQA~\cite{schwenk2022aokvqa}, and VSR~\cite{liu2023vsr}.
  \item \textbf{OCR Rater:} Optimizes for performance on TextVQA~\cite{singh2019textvqa}, ChartQA~\cite{masry2022chartqa}, and RD-Tablebench~\cite{reducto2024rdtablebench}.
  \item \textbf{STEM \& Knowledge Rater:} Optimizes for performance on M3Exam~\cite{zhang2023m3exam}, MathVista~\cite{lu2024mathvista}, and AI2D~\cite{kembhavi2016ai2d}.
\end{itemize}

Each rater is a $\sim$50M-parameter transformer operating on content features from a shared SigLIP 400M encoder \cite{siglip2} (see Appendix~\ref{app:rater_arch} for details).

\paragraph{Orthogonal Quality Signals.}
Training raters on distinct capability objectives should produce quality scores that capture genuinely different aspects of sample utility. If the resulting signals are low correlation and select for complementary subpopulations, this would enable independent data selection levers across capability domains and explain why decomposed raters outperform a monolithic scorer. We validate this in Section~\ref{sec:analysis}: the learned raters exhibit near-zero pairwise correlation and an effective dimensionality of 2.99/3.0, confirming that the decomposition yields orthogonal filtering dimensions rather than redundant structure.

\subsection{Curriculum Learning}
The original DataRater approach applies static top-$k$ filtering, training only on the highest-scored samples. We find that VLM training is more sensitive to early loss of visual diversity, likely because visual understanding benefits from exposure to a wide range of image types and compositions. We instead adopt a curriculum that progressively tightens quality thresholds over training by scheduling the effective fraction of retained data as a quadratic decay:
\begin{equation}
  E(t) \;=\; 1 - \left(\frac{t-1}{T}\right)^{2}, \qquad t \in \{1,\dots,T\},
  \label{eq:curriculum}
\end{equation}
where $T{=}10$ stages are spaced at 10k-step intervals. A sample $z$ is retained at stage $t$ if \emph{any} rater selects it (union rule):
\begin{equation}
  \texttt{retain}(z, t) = \bigvee_{c=1}^{C} \bigl[ r_{\phi_c}(z) \geq \tau_c(t) \bigr],
  \label{eq:union_rule}
\end{equation}
where $\tau_c(t)$ is the score threshold for rater $c$ at stage $t$. Because the raters produce near-orthogonal scores (Section~\ref{sec:analysis}), the union of $C$ independent top-$p$ selections retains approximately $1-(1-p)^{C}$ of the data; per-rater thresholds are obtained by inverting this relationship for each stage's target $E(t)$. This front-loads diversity: over 83\% of the corpus remains at stage 5, while the final stages concentrate on the top ${\sim}19\%$ of samples for capability refinement.

\begin{table}[t]
  \centering
  \small
  \caption{Curriculum schedule. Per-rater thresholds are set via $p(t)=1-\bigl(\frac{t-1}{T}\bigr)^{2/3}$; effective retention is the union across all three skill raters.}
  \vspace{3pt}
  \label{tab:curriculum_schedule}
  \begin{tabular}{ccccc}
    \toprule
    Stage & Per-Rater Top-$k$ & Samples Retained & Effective \% & Steps \\
    \midrule
     1 & 100.0\% & 6.14M & 100.0\% &  10k \\
     2 &  83.9\% & 6.12M &  99.7\% &  10k \\
     3 &  65.1\% & 5.87M &  95.7\% &  10k \\
     4 &  53.6\% & 5.54M &  90.3\% &  10k \\
     5 &  44.3\% & 5.12M &  83.4\% &  10k \\
     6 &  35.8\% & 4.57M &  74.6\% &  10k \\
     7 &  28.2\% & 3.94M &  64.3\% &  10k \\
     8 &  21.0\% & 3.16M &  51.5\% &  10k \\
     9 &  13.7\% & 2.19M &  35.6\% &  10k \\
    10 &   6.7\% & 1.16M &  19.0\% &  10k \\
    \bottomrule
  \end{tabular}
\end{table}

\subsection{Implementation}
The original DataRater was designed for text-only pretraining. Extending it to vision-language mid-training requires two adaptations.

\paragraph{Multimodal featurization.} Training data is represented as interleaved text and vision tokens in a single sequence. We extract content representations from a frozen SigLIP 400M encoder \cite{siglip2}, fuse them via a learned MLP, and pass them to the rater's transformer backbone \cite{perceptronai_isaac01_paper}.

\paragraph{Memory-efficient meta-learning.} The exact meta-gradient differentiates through the inner optimization trajectory, which introduces expensive second-order derivative structure and heavy memory pressure from storing or recomputing inner-loop intermediates. DataRater manages this cost with specialized gradient computation techniques \cite{kemaev2025mixflow}; absent these, the overhead scales prohibitively with model size and inner-loop length. We instead adopt a first-order surrogate with stop-gradient on the inner trajectory, eliminating second-order derivative terms at the cost of a biased gradient estimate. For capability $c$, the exact meta-gradient is
\begin{equation}
\eqbg{eqParamBg}{g_{\phi_c}^{\mathrm{SO}}}
=
\underbrace{
\nabla_{\eqbg{eqParamBg}{\phi_c}}
\eqbg{eqObjBg}{\mathcal{L}_{\mathrm{val}}^{(c)}}\!\left(\eqbg{eqStateBg}{\theta^{(S)}(\phi_c)}\right)
}_{\textcolor{eqLabelPurple}{\text{exact meta-gradient}}}
=
\underbrace{
\frac{\partial \eqbg{eqObjBg}{\mathcal{L}_{\mathrm{val}}^{(c)}}}{\partial \eqbg{eqStateBg}{\theta^{(S)}}}
\frac{\partial \eqbg{eqStateBg}{\theta^{(S)}}}{\partial \eqbg{eqParamBg}{\phi_c}}
}_{\textcolor{eqLabelBlue}{\text{through inner trajectory}}},
\end{equation}
which requires differentiating through all inner updates. We use a first-order surrogate with stop-gradient on the inner trajectory:
\begin{equation}
\tilde{\mathcal{L}}_{\mathrm{meta}}^{(c)}(\phi_c)
=
\frac{1}{K'}\sum_{k=1}^{K'}
\mathcal{L}_{\mathrm{val}}^{(c)}\!\left(
\operatorname{sg}\!\left[\theta^{(S)}\right],
\mathcal{V}_k;
r_{\phi_c}
\right),
\end{equation}
\begin{equation}
\eqbg{eqParamBg}{g_{\phi_c}^{\mathrm{FO}}}
=
\underbrace{
\nabla_{\eqbg{eqParamBg}{\phi_c}}
\tilde{\mathcal{L}}_{\mathrm{meta}}^{(c)}(\phi_c)
}_{\textcolor{eqLabelGreen}{\text{first-order surrogate}}},
\end{equation}
We also stream $\eqbg{eqDataBg}{K}$ train microbatches and accumulate inner gradients,
\begin{equation}
\underbrace{\eqbg{eqStateBg}{g_{\theta}^{(s)}}}_{\textcolor{eqLabelBlue}{\text{accumulated inner gradient}}}
=\sum_{k=1}^{\eqbg{eqDataBg}{K}}\nabla_{\eqbg{eqStateBg}{\theta^{(s)}}}\bar{\ell}_{k}^{(s)},
\qquad
\eqbg{eqStateBg}{\theta^{(s+1)}}=\eqbg{eqStateBg}{\theta^{(s)}}-
\underbrace{\frac{\eqbg{eqStepBg}{\alpha}}{\eqbg{eqDataBg}{K}}}_{\textcolor{eqLabelPurple}{\text{scaled step size}}}
\eqbg{eqStateBg}{g_{\theta}^{(s)}},
\qquad \eqbg{eqDataBg}{K\ge 1},
\end{equation}
which reduces peak activation memory from
\begin{equation}
\underbrace{\eqbg{eqObjBg}{M_{\mathrm{full}}=\Theta(BLH)}}_{\textcolor{eqLabelOrange}{\text{full-batch activations}}}
\quad\text{to}\quad
\underbrace{\eqbg{eqObjBg}{M_{\mathrm{stream}}=\Theta(B_{\mu}LH)}}_{\textcolor{eqLabelGreen}{\text{streamed activations}}},
\qquad \eqbg{eqDataBg}{B_{\mu}=B/K},
\end{equation}
with $B$ the effective batch size, $B_{\mu}$ the microbatch size, $L$ sequence length, and $H$ hidden width.

\begin{algorithm}[t]
\caption{SkillRater Training Loop (per rater)}
\label{alg:skillrater_loop}
\small
\begin{algorithmic}[1]
\State Initialize model parameters $\theta$ and rater parameters $\phi_c$
\For{$t=1,\dots,T$}
  \State $\theta^{(0)} \gets \theta$
  \State Sample $K$ train microbatches $\{\mathcal{B}_k\}_{k=1}^{K} \sim \mathcal{D}_{\mathrm{train}}$ \Comment{shared training pool}
  \State Compute weights $w_z\gets r_{\phi_c}(z)$ for all $z\in \bigcup_{k} \mathcal{B}_k$
  \For{$s=0,\dots,S-1$}
    \For{$k=1,\dots,K$}
      \State $\bar{\ell}_{k}^{(s)} \gets \frac{1}{|\mathcal{B}_k|}\sum_{z\in\mathcal{B}_k} w_z\,\ell(\theta^{(s)};z)$
    \EndFor
    \State $\theta^{(s+1)} \gets \theta^{(s)} - \alpha \nabla_{\theta^{(s)}} \bigl(\frac{1}{K}\sum_{k=1}^{K}\bar{\ell}_{k}^{(s)}\bigr)$
  \EndFor
  \State Sample $K'$ validation microbatches $\{\mathcal{V}_k\}_{k=1}^{K'} \sim \mathcal{D}_{\mathrm{val}}^{(c)}$ \Comment{capability-specific}
  \State $g_{\phi_c} \gets \nabla_{\phi_c}\tilde{\mathcal{L}}_{\mathrm{meta}}^{(c)}(\phi_c)$ \Comment{$\theta^{(S)}$ detached, gradient through rater path}
  \State $\phi_c \gets \phi_c - \eta\, g_{\phi_c}$
\EndFor
\end{algorithmic}
\end{algorithm}

\section{Results}

We evaluate \systemname on held out benchmarks and isolate the contribution of each component. We first report the primary held out comparison against unfiltered training and monolithic DataRater. We then compare curriculum scheduling against static filtering and evaluate transfer from 1B rater training to 2B model training. Together, these results show that capability specific decomposition and curriculum based data selection both contribute to the final gains.

\subsection{Held-Out Evaluation}
\label{sec:held_out}

\paragraph{Setup.}
We trained three capability-specific raters (Visual Understanding, OCR, and STEM \& Knowledge), each optimized on a disjoint set of training benchmarks. We then evaluated the resulting models on strictly held-out benchmarks that were never observed during any stage of rater optimization. Table~\ref{tab:held_out_split} shows the benchmark assignment for each category. All runs use the same Isaac 0.2 training recipe at the 2B parameter scale and train on Mammoth \cite{guo2024mammothvlelicitingmultimodalreasoning}. To ensure fair comparison, both the DataRater and \systemname Combined baselines use 150M total rater parameters and the same curriculum schedule (Eq.~\eqref{eq:curriculum}) as the final \systemname configuration; the only difference is how the rater capacity is allocated (one monolithic rater vs.\ three capability-specific raters).

\begin{table}[t]
  \centering
  \small
  \setlength{\tabcolsep}{5pt}
  \renewcommand{\arraystretch}{1.12}
  \caption{Held-out benchmark split by capability. Training benchmarks supervise rater learning, and held-out benchmarks are used only for final evaluation.}
  \vspace{3pt}
  \label{tab:held_out_split}
  \begin{tabular}{@{}l >{\raggedright\arraybackslash}p{0.34\linewidth} >{\raggedright\arraybackslash}p{0.34\linewidth}@{}}
    \toprule
    \textbf{Capability} & \textbf{Rater Training Benchmarks} & \textbf{Held-Out Evaluation Benchmarks} \\
    \midrule
    Visual Understanding & SEED-Bench~\cite{li2023seedbench}, A-OKVQA~\cite{schwenk2022aokvqa}, VSR~\cite{liu2023vsr} & VQAv2~\cite{goyal2017makingVQA}, NLVR2~\cite{suhr2019nlvr2}, MME~\cite{fu2023mme} \\
    OCR & TextVQA~\cite{singh2019textvqa}, ChartQA~\cite{masry2022chartqa}, RD-Tablebench~\cite{reducto2024rdtablebench} & DocVQA~\cite{mathew2021docvqa}, ChartMuseum~\cite{tang2025chartmuseum}, \mbox{OCRBenchV2~\cite{fu2024ocrbenchv2}} \\
    STEM \& Knowledge & M3Exam~\cite{zhang2023m3exam}, MathVista~\cite{lu2024mathvista}, AI2D~\cite{kembhavi2016ai2d} & MathVision~\cite{wang2024mathvision}, MMMU~\cite{yue2024mmmu_cvpr}, MMMU-Pro~\cite{yue2025mmmupro_acl} \\
    \bottomrule
  \end{tabular}
\end{table}

\paragraph{Results.}
Table~\ref{tab:held_out_results} reports held-out performance at 100k steps (full results at 25k/50k/75k/100k are provided in Appendix~\ref{sec:appendix_full_results}). The monolithic DataRater baseline improves overall accuracy by 1.21 points over unfiltered training (45.89\% vs.\ 44.68\%), but still underperforms \systemname on OCR and STEM (48.47\% vs.\ 50.05\% on OCR; 20.88\% vs.\ 23.61\% on STEM \& Knowledge). A pooled-validation baseline improves overall performance to 47.01\% (\systemname Combined: a single 150M rater trained on the union of all validation tasks), and full \systemname further improves to 48.40\%. The gains from \systemname Combined to full \systemname are largest on OCR and STEM, indicating that using a union rule over capability-specific raters adds value beyond a single pooled rater and curriculum alone.

\begin{table}[t]
  \centering
  \small
  \caption{Held-out evaluation at 100k steps. All benchmarks were never seen during rater training.}
  \vspace{3pt}
  \label{tab:held_out_results}
  \begin{tabular}{lcccc}
    \toprule
    Category & Mammoth (Baseline) & DataRater & \systemname Combined & \systemname \\
    \midrule
    STEM \& Knowledge & 20.08\% & 20.88\% & 22.39\% & \textbf{23.61\%} \\
    OCR & 48.05 \% & 48.47\% & 47.28\%  & \textbf{50.05\%} \\
    Visual Understanding & 65.91\% & 68.33\% & 71.36\%  & \textbf{71.54\%} \\
    \midrule
    Overall & 44.68\% & 45.89\% &  47.01\% & \textbf{48.40\%} \\
    \bottomrule
  \end{tabular}
\end{table}
\textit{Overall} is computed as the unweighted mean of the three capability categories.

\subsection{Curriculum vs Static Filtering}
Table~\ref{tab:filtering_comparison} compares curriculum learning against static top-$k$\% filtering at 100k training steps. In this setting, we use a single 50M rater trained for a single capability, and we apply its scores to filter the mid-training data stream. The static results show that no single threshold dominates across benchmarks: top 50\% gives the best VQAv2 score (75.7), top 90\% leads on DocVQA (84.6) and TextVQA (69.8), and top 80\% is best on RealWorldQA (59.8). This dispersion suggests that different capabilities prefer different quality-diversity tradeoffs, so a fixed threshold is suboptimal.

Our curriculum schedule resolves these conflicting optima by tightening the threshold over training following a quadratic decay $E(t)=1-(t{-}1)^2/T^2$ (Eq.~\eqref{eq:curriculum}), starting from unfiltered data and progressively compressing to the top ${\sim}19\%$ by stage 10. Early training retains broad coverage to build general visual representations, while later training concentrates on high-scoring samples to refine capability performance. Curriculum achieves the best overall average (64.1\%), exceeding the best static threshold (top 90\% at 63.7\%).

\begin{table*}[t]
\centering
\small
\setlength{\tabcolsep}{4pt}
\renewcommand{\arraystretch}{1.08}
\caption{Filtering strategy comparison at 100k steps on VQA benchmarks. Best result per column is in bold.}
\vspace{3pt}
\label{tab:filtering_comparison}
\resizebox{\linewidth}{!}{%
\begin{tabular}{
  l
  r r r r r r r r r r
}
\toprule
Method & {VQAv2\cite{goyal2017makingVQA}} & {DocVQA\cite{mathew2021docvqa}} & {RWQA\cite{xai2024realworldqa}} & {TextVQA\cite{singh2019textvqa}} & {OCRBench\cite{fu2024ocrbenchv2}} & {MME\cite{fu2023mme}} & {SEED\cite{li2023seedbench}}& {BLINK\cite{fu2024blink}} & {VSR\cite{liu2023vsr}} & {Avg.} \\
\midrule
\rowcolor{gray!12} Curriculum & 67.6 & 79.6 & 56.5 & 61.0 & 79.1 & \textbf{76.6} & \textbf{53.8} & 34.2 & \textbf{68.6} & \textbf{64.1} \\
\midrule
Top-90\% & 74.6 & \textbf{84.6} & 58.6 & \textbf{69.8} & \textbf{80.9} & 59.4 & 51.0 & 30.4 & 63.8 & 63.7 \\
Top-60\% & 69.8 & 84.0 & 58.1 & 66.4 & 80.4 & 60.1 & 37.8 & 37.0 & 68.0 & 62.4 \\
Top-50\% & \textbf{75.7} & 83.1 & 57.9 & 66.2 & 80.7 & 58.5 & 44.4 & \textbf{40.6} & 60.0 & 63.0 \\
Top-80\% & 60.6 & 82.4 & \textbf{59.8} & 59.0 & 80.1 & 57.4 & 47.8 & 32.8 & 65.4 & 60.6 \\
Top-70\% & 70.1 & 83.1 & 57.7 & 64.3 & 81.5 & 60.0 & 45.2 & 28.6 & 68.4 & 62.1 \\
Top-40\% & 71.7 & 80.7 & 58.2 & 69.6 & 80.4 & 55.8 & 53.2 & 34.0 & 64.0 & 63.1 \\
Top-30\% & 71.9 & 78.2 & 56.2 & 63.8 & 79.4 & 58.8 & 52.2 & 37.0 & 52.6 & 61.1 \\
Top-20\% & 68.0 & 77.2 & 52.6 & 65.4 & 76.9 & 53.5 & 45.4 & 33.8 & 48.0 & 57.9 \\
Top-10\% & 55.6 & 61.5 & 54.1 & 52.4 & 60.2 & 52.9 & 46.2 & 29.6 & 53.8 & 51.8 \\
\bottomrule
\end{tabular}}
\end{table*}

\subsection{Scale Transfer}
\begin{figure}[t]
  \centering
  \includegraphics[width=0.95\linewidth]{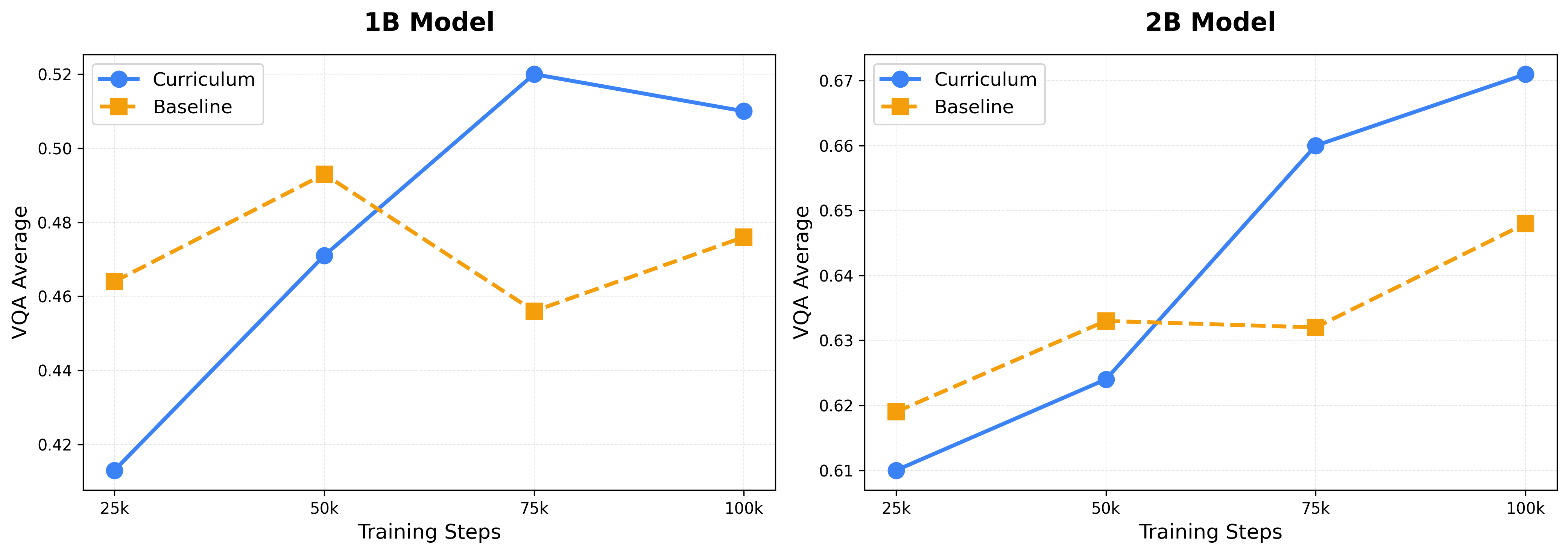}
  \caption{Scale transfer at 1B and 2B parameters. Raters trained at 1B improve both scales without retraining, and the curriculum gap over unfiltered training widens over steps.}
  \label{fig:scaling_600m}
\end{figure}

Figure~\ref{fig:scaling_600m} shows training dynamics at 1B and 2B parameters. The capability-specific raters are trained at 1B and transferred to 2B without retraining. At both scales, curriculum with \systemname outperforms the unfiltered baseline, and the gap increases over training. At the 2B held-out endpoint (Table~\ref{tab:held_out_results}), this corresponds to a +3.72 point overall gain over unfiltered training (48.40\% vs.\ 44.68\%).

Across held-out evaluation, curriculum ablations, and scale transfer, the result is consistent. Capability-specific decomposition improves the data signal, and curriculum over their union converts that signal into larger downstream gains.

\FloatBarrier
\section{Analysis}
\label{sec:analysis}

The held-out results show that decomposition and curriculum improve performance. Two questions remain: are the learned rater signals genuinely independent, and what distinguishes retained from filtered samples?

\begin{figure}[t]
  \centering
  \includegraphics[width=0.95\linewidth]{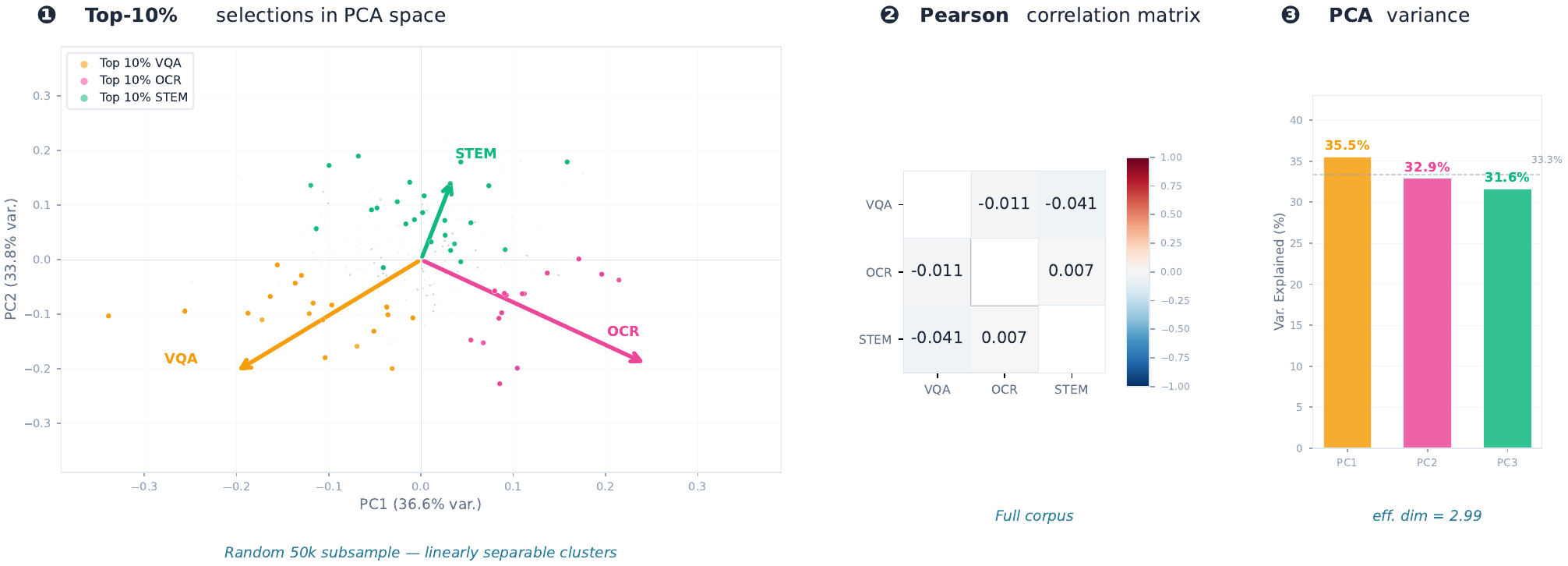}
  \caption{Orthogonality of capability-specific rater signals. Top-scoring subsets from each rater align with distinct directions in score space, indicating low overlap and complementary selection.}
  \label{fig:orthogonality}
\end{figure}

\paragraph{Orthogonality.}
Multimodal quality in our setting is not a single axis. If capability-specific raters learned redundant signals, decomposition would add little over a monolithic scorer. The learned signals are not redundant. Across the full training corpus, pairwise Pearson correlations range from $-0.011$ (VQA--OCR) to $0.007$ (OCR--STEM), with mean absolute correlation $0.020$. Spearman rank correlations are similarly low ($-0.004$, $-0.045$, $0.007$; mean $0.019$). Principal component analysis shows near-uniform variance across components (35.5\%, 32.9\%, 31.6\%), yielding effective dimensionality 2.99 out of 3.0. Figure~\ref{fig:orthogonality} visualizes this on 50,000 random samples: top-10\% selections from each rater align with distinct, linearly separable directions in score space.

This structure explains the held-out gap over monolithic filtering (Table~\ref{tab:held_out_results}). DataRater improves overall accuracy by 1.21 points, but it compresses low-correlation capability signals into one scalar and yields smaller gains on OCR and STEM than capability-specific decomposition. Decomposed raters avoid that compression because each rater optimizes its own axis. The curriculum union rule then retains complementary high-value samples rather than repeatedly selecting the same subset.

\begin{figure}[t]
  \centering
  \includegraphics[width=0.95\linewidth]{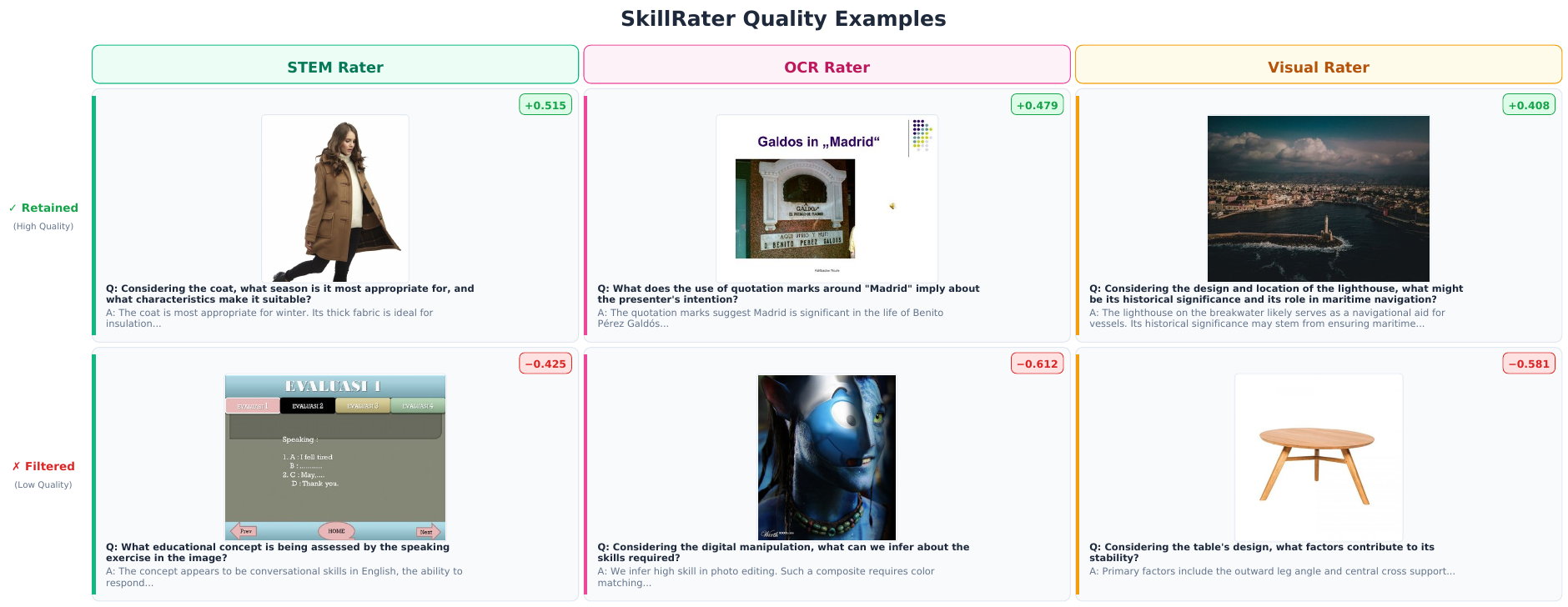}
  \caption{Qualitative retained and filtered samples for each capability-specific rater. Retained examples demand capability-aligned reasoning, while filtered examples are shallow or mismatched for that capability.}
  \label{fig:qualitative_examples}
\end{figure}

\paragraph{Qualitative Rating Examples.}
Figure~\ref{fig:qualitative_examples} shows representative retained and filtered samples for each capability-specific rater. For STEM, retained samples require multi-step reasoning about physical properties, while filtered samples are shallow recognition tasks. For OCR, retained samples require reading and interpreting embedded text in context, while filtered samples do not use text meaningfully. For Visual Understanding, retained samples require grounded visual reasoning, such as inferring a lighthouse's design and strategic harbor placement, while filtered samples are generic structural descriptions. Across all raters, retained samples align with capability-specific reasoning demands.

\section{Related Work}
\label{sec:related_work}

Data curation methods for foundation models generally follow three lines: heuristic filtering, learned scoring, and capability aligned selection. \systemname differs from prior work in both mechanism and setting. We replace a single quality score with a mixture of capability specific raters, and we study multimodal mid training with interleaved image--text data.

\paragraph{Heuristic and Unsupervised Filtering.}
Heuristic filters dominate large scale multimodal curation. CLIP score filtering~\cite{radford2021clip, schuhmann2022laion5b, gadre2024datacomp} retains pairs with high image--text similarity, aesthetic scoring~\cite{schuhmann2022laion5b} predicts visual appeal from learned embeddings, and deduplication~\cite{lee2022dedup} removes near duplicates through hashing. These methods are efficient and robust at web scale, but each fixes quality to one axis. In practice this misses examples that are weak on surface alignment or aesthetics but strong for capability specific learning signals such as OCR or structured reasoning.

\paragraph{Learned Data Selection.}
Learned selectors replace fixed rules with supervision from target corpora, training dynamics, or validation objectives. DFN~\cite{fang2024dfn} trains a classifier from curated partitions for CLIP pre training. DSIR~\cite{xie2023dsir} and DoReMi~\cite{xie2023doremi} reweight text by importance resampling and robust domain optimization. EL2N and GraNd~\cite{paul2021el2n} prune by early training statistics, LESS~\cite{xia2024less} and Rho-1~\cite{lin2024rho1} select data by gradient influence or token level loss gaps, and AutoMixer~\cite{chang2025automixer} leverages checkpoint artifacts as capability aware data mixers via first order influence approximation.
DataRater~\cite{calian2025datarater}, which we build on, learns scores with bilevel meta learning and has strong text only results. The shared limitation is scalar compression: these methods output one score per sample, so conflicting capability signals are merged before selection. Our held out results show this bottleneck directly, where monolithic DataRater yields limited gains and an OCR regression relative to unfiltered training (Table~\ref{tab:held_out_results}).

\paragraph{Capability Driven Selection.}
Recent work aligns selection to downstream capabilities instead of generic quality. IFD~\cite{li2024cherry}, Skill-Mix~\cite{yu2023skillmix}, MoS~\cite{wu2024mos}, and AlpaGasus~\cite{chen2023alpagasus} all target instruction data utility, supporting the view that quality is multi dimensional and task dependent. BETR~\cite{mizrahi2025betr} extends this thesis to pre training with benchmark targeted ranking, reporting a $2.1\times$ compute multiplier over strong baselines. These results are consistent with our motivation, but existing methods remain text only or apply a single targeting channel. \systemname extends capability aligned selection to multimodal mid training by learning orthogonal raters through meta learning and composing them with a curriculum over the union of rater scores.

\section{Conclusion \& Future Work}
Monolithic scalar scoring creates capability tradeoffs in multimodal mid training. A single DataRater provides modest overall gains but regresses on OCR relative to unfiltered training. \systemname addresses this by decomposing filtering into capability aligned raters trained via meta learning, combined through a curriculum that preserves diversity early and increases selectivity later.

On held out benchmarks, \systemname improves over unfiltered training by 5.63\% on visual understanding, 2.00\% on OCR, and 3.53\% on STEM, with a +3.93 point overall gain. Rater scores are near orthogonal (mean pairwise correlation 0.020, effective dimensionality 2.99/3.0), and curriculum over the rater union outperforms static top $k$ filtering. Raters trained at 1B transfer to 2B without retraining, indicating that the learned signals generalize across scale.

Several directions remain open. First, we use three capability dimensions; understanding how many orthogonal raters the framework supports before diminishing returns is an empirical question with implications for how finely capabilities should be decomposed. Second, the curriculum schedule is manually designed. Jointly learning the raters and the schedule through bilevel optimization over curriculum parameters is a natural extension. Third, rater quality is bounded by benchmark coverage: capabilities not represented in the validation set cannot be targeted. Expanding to broader capability taxonomies or learning raters from weaker supervision signals would reduce this dependency.

More broadly, the near orthogonality of learned rater signals supports the view that data quality is intrinsically multidimensional. This property is unlikely to be specific to multimodal mid training. Text pretraining, code generation, and instruction tuning all involve competing capability demands, and scalar quality scores may face similar compression bottlenecks in those settings. Capability aligned decomposition offers a principled alternative.

\bibliographystyle{unsrtnat}
\bibliography{references}

\clearpage

\appendix
\section{Implementation Details}
\label{sec:appendix_impl}

\subsection{DataRater hyperparameters}
\paragraph{Hyperparameters.}
We use $S=2$ inner steps, accumulate over $K=K'=32$ microbatches (effective batch size 128), and update the rater every 250 training steps.

\subsection{Rater architecture details}
\label{app:rater_arch}
All raters share the same architecture: a 6-layer transformer with 768 hidden dimensions and 12 attention heads ($\sim$50M parameters), operating on content features from a shared SigLIP 400M encoder. Scores are normalized with a sigmoid using temperature $\tau=0.1$. We apply entropy regularization (coefficient 0.2) to prevent score collapse while maintaining sufficient weight variance.

\subsection{Benchmark trajectories across training}
\label{sec:appendix_trajectories}

\begin{table}[H]
  \centering
  \scriptsize
  \setlength{\tabcolsep}{2.2pt}
  \renewcommand{\arraystretch}{1.10}
  \caption{VQA benchmark trajectories at 25k/50k/75k/100k steps.}
  \label{tab:appendix_vqa_trajectories}
  \resizebox{\linewidth}{!}{%
  \begin{tabular}{lcccccccccccc}
    \toprule
    & \multicolumn{4}{c}{SkillRater}
    & \multicolumn{4}{c}{Baseline}
    & \multicolumn{4}{c}{Original DataRater} \\
    \cmidrule(lr){2-5} \cmidrule(lr){6-9} \cmidrule(lr){10-13}
    Benchmark & 25k & 50k & 75k & 100k & 25k & 50k & 75k & 100k & 25k & 50k & 75k & 100k \\
    \midrule
    VQAv2       & 72.24 & 72.46 & 71.88 & 74.01 & 71.81 & 67.84 & 65.57 & 70.15 & 69.64 & 72.96 & 67.92 & 66.70 \\
    RealWorldQA & 53.53 & 58.12 & 54.97 & 61.26 & 60.60 & 60.34 & 61.26 & 60.47 & 53.01 & 57.20 & 55.76 & 57.72 \\
    TextVQA     & 57.85 & 42.20 & 64.49 & 67.07 & 61.44 & 66.75 & 65.66 & 69.45 & 48.49 & 57.02 & 56.74 & 57.91 \\
    MME         & 62.68 & 63.49 & 67.86 & 67.73 & 55.19 & 56.10 & 57.87 & 58.43 & 62.80 & 66.86 & 66.21 & 64.84 \\
    SEED-Bench  & 49.40 & 53.60 & 61.00 & 56.80 & 50.80 & 50.60 & 48.80 & 50.80 & 44.60 & 52.80 & 50.00 & 53.00 \\
    BLINK       & 31.40 & 33.40 & 36.80 & 31.60 & 33.60 & 34.60 & 33.40 & 33.40 & 32.00 & 34.40 & 32.40 & 30.20 \\
    VSR         & 63.20 & 66.20 & 67.60 & 71.60 & 61.60 & 57.00 & 57.40 & 56.00 & 66.40 & 60.40 & 65.60 & 69.80 \\
    \bottomrule
  \end{tabular}}
\end{table}

\section{Additional Results}
\label{sec:appendix_full_results}

\begin{table}[ht]
\centering
\scriptsize
\setlength{\tabcolsep}{4.5pt}
\caption{Category average results at 25k, 50k, 75k, and 100k steps.}
\label{tab:appendix_category_averages}
\resizebox{\linewidth}{!}{%
\begin{tabular}{llcccc}
\toprule
\textbf{Step} & \textbf{Category} & \textbf{Mammoth} & \textbf{DataRater} & \textbf{Combined} & \textbf{\systemname} \\
\midrule
\multirow{4}{*}{25k}
  & STEM \& Knowledge & 19.13 & 20.90 & \textbf{22.48} & 21.25 \\
  & OCR & 43.47 & \textbf{44.84} & 42.78 & 44.59 \\
  & Visual Understanding & 67.18 & 67.78 & \textbf{68.95} & 68.53 \\
  & \textbf{Overall} & 43.26 & 44.51 & 44.74 & \textbf{44.79} \\
\midrule
\multirow{4}{*}{50k}
  & STEM \& Knowledge & 17.45 & 20.16 & 20.59 & \textbf{20.61} \\
  & OCR & 46.56 & 46.24 & 45.70 & \textbf{46.87} \\
  & Visual Understanding & 66.63 & 68.51 & \textbf{70.67} & 68.76 \\
  & \textbf{Overall} & 43.55 & 44.97 & \textbf{45.65} & 45.41 \\
\midrule
\multirow{4}{*}{75k}
  & STEM \& Knowledge & 19.15 & 21.24 & \textbf{22.86} & 22.58 \\
  & OCR & 46.82 & 45.81 & 47.89 & \textbf{48.15} \\
  & Visual Understanding & 65.43 & 68.27 & 70.73 & \textbf{70.74} \\
  & \textbf{Overall} & 43.80 & 45.11 & \textbf{47.16} & \textbf{47.16} \\
\midrule
\multirow{4}{*}{100k}
  & STEM \& Knowledge & 20.07 & 20.88 & 22.39 & \textbf{23.61} \\
  & OCR & 48.05 & 48.46 & 47.28 & \textbf{50.05} \\
  & Visual Understanding & 65.91 & 68.33 & 71.36 & \textbf{71.54} \\
  & \textbf{Overall} & 44.68 & 45.89 & 47.01 & \textbf{48.40} \\
\bottomrule
\end{tabular}}
\end{table}

\FloatBarrier

\end{document}